\documentclass[runningheads,a4paper]{llncs}

\usepackage{amssymb}
\usepackage{amsmath}
\usepackage{graphicx}
\usepackage{verbatim}
\usepackage{multirow}

\usepackage{url}
\urldef{\mailsa}\path|chuanqi.tan@gmail.com|
\urldef{\mailsb}\path|fcsun@tsinghua.edu.cn| 
 
\newcommand{\keywords}[1]{\par\addvspace\baselineskip
\noindent\keywordname\enspace\ignorespaces#1}

\begin{document}

\mainmatter  

\title{A Survey on Deep Transfer Learning}

\titlerunning{A Survey on Deep Transfer Learning}

\author{Chuanqi Tan$^1$ \and
    Fuchun Sun$^2$
\and Tao Kong$^1$
\and \\Wenchang Zhang$^1$
\and Chao Yang$^1$
\and Chunfang Liu$^2$
}
\authorrunning{Chuanqi Tan et al.}

\institute{State Key Laboratory of Intelligent Technology and Systems
    \\Tsinghua National Laboratory for Information Science and Technology (TNList)
\\Department of Computer Science and Technology, Tsinghua University
\\$^1$\{tcq15, kt14, zhangwc14, yang-c15\}@mails.tsinghua.edu.cn
\\$^2$\{fcsun, cfliu1985\}@tsinghua.edu.cn
}

\toctitle{Lecture Notes in Computer Science}
\tocauthor{Authors' Instructions}
\maketitle

\begin{abstract}
As a new classification platform, deep learning has recently received increasing attention from researchers and has been successfully applied to many domains. 
In some domains, like bioinformatics and robotics, it is very difficult to construct a large-scale well-annotated dataset due to the expense of data acquisition and costly annotation, which limits its development.
Transfer learning relaxes the hypothesis that the training data must be independent and identically distributed (i.i.d.) with the test data, which motivates us to use transfer learning to solve the problem of insufficient training data.
This survey focuses on reviewing the current researches of transfer learning by using deep neural network and its applications. We defined deep transfer learning, category and review the recent research works based on the techniques used in deep transfer learning. 

\end{abstract}

\section{Introduction}

Deep learning has recently received increasing attention from researchers and has been successfully applied to numerous real-world applications. 
Deep learning algorithms attempt to learn high-level features from mass data, which make deep learning beyond traditional machine learning. 
It can automatic extract data features by unsupervised or semi-supervised feature learning algorithm and hierarchical feature extraction. 
In contrast, traditional machine learning methods need to design features manually that seriously increases the burden on users.
It can be said that deep learning is an representation learning algorithm based on large-scale data in machine learning.

\textit{Data dependence} is one of the most serious problem in deep learning. Deep learning has a very strong dependence on massive training data compared to traditional machine learning methods,  because it need a large amount of data to understand the latent patterns of data.
An interesting phenomenon can be found that the scale of the model and the size of the required amount of data has a almost linear relationship.
An acceptable explanation is that for a particular problem, the expressive space of the model must be large enough to discover the patterns under the data . The pre-order layers in the model can identify high-level features of training data, and the subsequent layers can identify the information needed to help make the final decision.

\textit{Insufficient training data} is a inescapable problem in some special domains.
The collection of data is complex and expensive that make it is extremely difficult to build a large-scale, high-quality annotated dataset.
For example, each sample in bioinformatics dataset often demonstration a clinical trial or a painful patient. 
In addition, even we obtain training dataset by paid an expensive price, it is very easy to get out of date and thus cannot be effectively applied in the new tasks.

Transfer learning relaxes the hypothesis that the training data must be independent and identically distributed (i.i.d.) with the test data, which motivates us to use transfer learning to against the problem of insufficient training data. In transfer learning, the training data and test data are not required to be i.i.d., and the model in target domain is not need to trained from scratch, which can significantly reduce the demand of training data and training time in the target domain.

In the past, most studies of transfer learning were conducted in traditional machine learning methods. Due to the dominance position of deep learning in modern machine learning methods, a survey on deep transfer learning and its applications is particularly important.
The \textbf{contributions} of this survey paper are as follows:

\begin{itemize}

\item We define the deep transfer learning and categorizing it into four categories for the first time.

\item We reviewing the current research works on each category of deep transfer learning, and given a standardized description and sketch map of every category.

\end{itemize}

\section{Deep Transfer Learning}

\begin{figure}[h]
    \centering
    \setlength{\fboxsep}{0cm} 
    \includegraphics[width=2.5in]{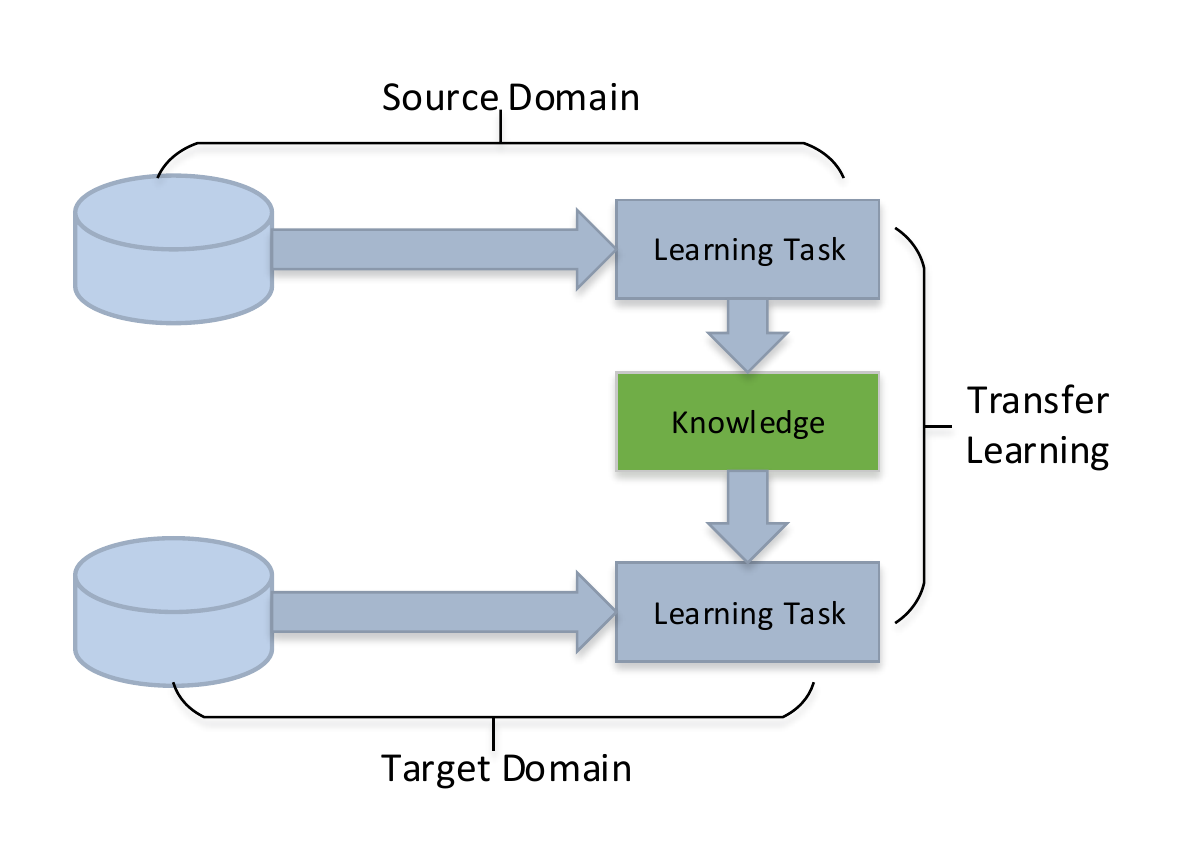}
    \caption{Learning process of transfer learning. }
    \label{learning process of transfer learning}
\end{figure}

Transfer learning is an important tool in machine learning to solve the basic problem of insufficient training data.
It try to transfer the knowledge from the source domain to the target domain by relaxing the assumption that the training data and the test data must be i.i.d. 
This will leads to a great positive effect on many domains that are difficult to improve because of insufficient training data. 
The learning process of transfer learning illustrated in the Fig. \ref{learning process of transfer learning}.

Some notations used in this survey need to be clearly defined. First of all, we give the definitions of a domain and a task respectively:
A domain can be represented by $\mathcal{D}=\{\chi,P(X)\}$, which contains two parts: the feature space $\chi$ and the edge probability distribution $P(X)$ where $X=\{x_1,...,x_n\} \in \chi$.
A task can be represented by $\mathcal{T}=\{y, f(x)\}$. It consists of two parts: label space $y$ and target prediction function $f(x)$. $f(x)$ can also be regarded as a conditional probability function $P(y|x)$.
Then, the transfer learning can be formal defined as follows:

\begin{definition} 
\textbf{(Transfer Learning).} Given a learning task $\mathcal{T}_t$ based on $\mathcal{D}_t$, and we can get the help from $\mathcal{D}_s$ for the learning task $\mathcal{T}_s$. Transfer learning aims to improve the performance of predictive function $f_{\mathcal{T}}(\cdot)$ for learning task $\mathcal{T}_t$ by discover and transfer latent knowledge from $\mathcal{D}_s$ and $\mathcal{T}_s$, where $\mathcal{D}_s \neq \mathcal{D}_t$ and/or $\mathcal{T}_s \neq \mathcal{T}_t$. In addition, in the most case, the size of $\mathcal{D}_s$ is much larger than the size of $\mathcal{D}_t$, $N_s \gg N_t$.
\end{definition}

Surveys \cite{pan2010survey} and \cite{weiss2016survey} divide the transfer learning methods into three major categories with the relationship between the source domain and the target domain, which has been widely accepted. 
These suverys are good summary of the past works on transfer learning, which introduced a number of classic transfer learning methods.
Further more, many newer and better methods have been proposed recently.
In recent years, transfer learning research community are mainly focused on the following two aspects: domain adaption and multi-source domains transfer.

Nowadays, deep learning has achieved dominating situation in many research fields in recent years. It is important to find how to effectively transfer knowledge by deep neural network, which called deep transfer learning that defined as follows:

\begin{definition} 
    \textbf{(Deep Transfer Learning).} Given a transfer learning task defined by $\langle\mathcal{D}_s, \mathcal{T}_s, \mathcal{D}_t, \mathcal{T}_t, f_{\mathcal{T}}(\cdot)\rangle$. It is a deep transfer learning task where $f_{\mathcal{T}}(\cdot)$ is a non-linear function that reflected a deep neural network. 
\end{definition}

\section{Categories}

Deep transfer learning studies how to utilize knowledge from other fields by deep neural networks. Since deep neural networks have become popular in various fields, a considerable amount of deep transfer learning methods have been proposed that it is very important to classify and summarize them.
Based on the techniques used in deep transfer learning, this paper classifies deep transfer learning into four categories: instances-based deep transfer learning, mapping-based deep transfer learning, network-based deep transfer learning, and adversarial-based deep transfer learning, which are shown in Table \ref{table: Categorizing of Deep Transfer Learning}.

\begin{table}[h]
    \centering
    \caption{Categorizing of deep transfer learning.}
    \label{table: Categorizing of Deep Transfer Learning}
    \begin{tabular}
        {p{0.22\columnwidth}|p{0.52\columnwidth}|p{0.24\columnwidth}}
        \hline\hline\noalign{\smallskip}
        Approach category & Brief description & Some related works\\ 
        
        \hline
        Instances-based & Utilize instances in source domain by appropriate weight.& 
        \cite{dai2007boosting}, \cite{yao2010boosting}, \cite{pardoe2010boosting}, \cite{wan2011bi}, \cite{li2017transfer}, \cite{xu2017unified}, \cite{liu2018ensemble}\\ 
        \hline
        Mapping-based & Mapping instances from two domains into a new data space with better similarity.  & 
        \cite{tzeng2014deep}, \cite{long2015learning}, \cite{gretton2012optimal}, \cite{long2016deep}, \cite{arjovsky2017wasserstein}  \\ 
        \hline
        Network-based & Reuse the partial of network pre-trained in the source domain. &
        \cite{huang2013cross}, \cite{oquab2014learning}, \cite{long2016unsupervised}, \cite{zhu2016deep}, \cite{chang2017unsupervised}, \cite{george2017deep}, \cite{yosinski2014transferable} \\ 
        \hline
        Adversarial-based & Use adversarial technology to find transferable features that both suitable for two domains.  &
        \cite{ajakan2014domain}, \cite{ganin2014unsupervised}, \cite{tzeng2015simultaneous}, \cite{tzeng2017adversarial}, \cite{long2017domain}, \cite{luo2017label} \\ 
        \hline\hline
    \end{tabular}
\end{table}

\subsection{Instances-based deep transfer learning}

\begin{figure}[h]
	\centering
	\setlength{\fboxsep}{0cm} 
	\includegraphics[page=1,trim=4cm 7cm 2cm 6.5cm,clip,width=0.9\textwidth]{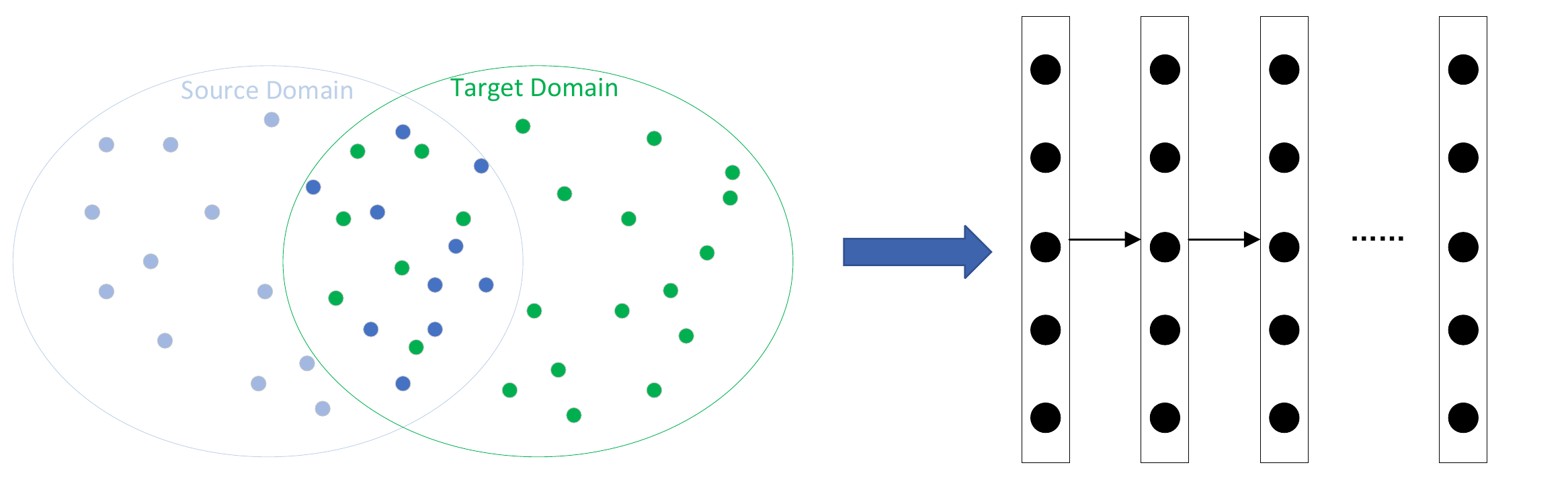}
	\caption{Sketch map of instances-based deep transfer learning. 
		Instances with light blue color in source domain meanings dissimilar with target domain are exclude from training dataset; Instances with dark blue color in source domain meanings similar with target domain are include in training dataset with appropriate weight. 
	}
	\label{Processing of instances-based deep transfer learning}
\end{figure}

Instances-based deep transfer learning refers to use a specific weight adjustment strategy, select partial instances from the source domain as supplements to the training set in the target domain by assigning appropriate weight values to these selected instances.
It is based on the assumption that "\textit{Although there are different between two domains, partial instances in the source domain can be utilized by the target domain with appropriate weights.}".
The sketch map of instances-based deep transfer learning are shown in Fig. \ref{Processing of instances-based deep transfer learning}.

TrAdaBoost proposed by \cite{dai2007boosting} use AdaBoost-based technology to filter out instances that are dissimilar to the target domain in source domains. Re-weighted instances in source domain to compose a distribution similar to target domain. Finally, training model by using the re-weighted instances from source domain and origin instances from target domain. It can reduce the weighted training error on different distribution domains that preserving the properties of AdaBoost.
TaskTrAdaBoost proposed by \cite{yao2010boosting} is a fast algorithm promote rapid retraining over new targets.
Unlike TrAdaBoost is designed for classification problems, ExpBoost.R2 and TrAdaBoost.R2 were proposed by \cite{pardoe2010boosting} to cover the regression problem.
Bi-weighting domain adaptation (BIW) proposed \cite{wan2011bi} can aligns the feature spaces of two domains into the common coordinate system, and then assign an appropriate weight of the instances from source domain.
\cite{li2017transfer} propose a enhanced TrAdaBoost to handle the problem of interregional sandstone microscopic image classification.
\cite{xu2017unified} propose a metric transfer learning framework to learn instance weights and a distance of two different domains in a parallel framework to make knowledge transfer across domains more effective.
\cite{liu2018ensemble} introduce an ensemble transfer learning to deep neural network that can utilize instances from source domain.

\subsection{Mapping-based deep transfer learning}

\begin{figure}[h]
	\centering
	\setlength{\fboxsep}{0cm} 
	\includegraphics[page=2,trim=3.9cm 4.6cm 2.8cm 5.6cm,clip,width=0.9\textwidth]{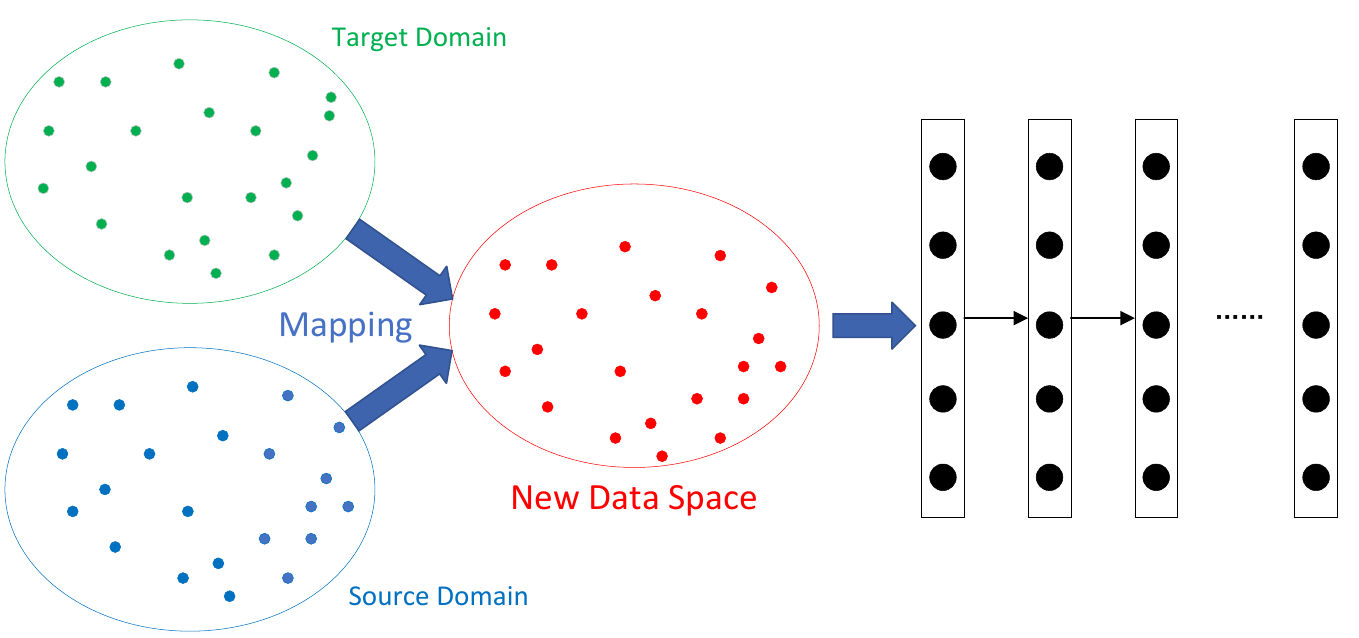}
	\caption{Sketch map of mapping-based deep transfer learning. Simultaneously, instances from source domain and target domain are mapping to a new data space with more similarly. Consider all instances in the new data space as the training set of the neural network.}
	\label{Processing of Mapping-based deep transfer learning}
\end{figure}

Mapping-based deep transfer learning refers to mapping instances from the source domain and target domain into a new data space. In this new data space, instances from two domains are similarly and suitable for a union deep neural network.
It is based on the assumption that "\textit{Although there are different between two origin domains, they can be more similarly in an elaborate new data space.}".
The sketch map of instances-based deep transfer learning are shown in Fig. \ref{Processing of Mapping-based deep transfer learning}.

Transfer component analysis (TCA) introduced by \cite{Pan2011Domain} and TCA-based methods \cite{Zhang2017Joint} had been widely used in many applications of traditional transfer learning. A natural idea is extend the TCA method to deep neural network.
\cite{tzeng2014deep} extend MMD to comparing distributions in a deep neural network, by introduces an adaptation layer and an additional domain confusion loss to learn a representation that is both semantically meaningful and domain invariant. 
The MMD distance used in this work is defined as
\begin{equation}
D_{\mathcal{MMD}}(X_S, X_T) =\left|\left|
\dfrac{1}{|X_S|} \sum_{x_s \in X_S}\phi(x_s) - \dfrac{1}{|X_T|} \sum_{x_t \in X_T}\phi(x_t)
\right|\right|
\end{equation}
and the loss function is defined as
\begin{equation}
\mathcal{L} = \mathcal{L}_C(X_L, y) + \lambda D^2_{\mathcal{MMD}}(X_S, X_T).
\end{equation}
\cite{long2015learning} improved previous work by replace MMD distance with multiple kernel variant MMD (MK-MMD) distance proposed by \cite{gretton2012optimal}.
The hidden layer related with the learning task in the convolutional neural networks (CNN) is mapped into the reproducing kernel Hilbert space (RKHS), and the distance between different domains is minimized by the multi-core optimization method.
\cite{long2016deep} propose joint maximum mean discrepancy (JMMD) to measurement the relationship of joint distribution. JMMD was used to generalize the transfer learning ability of the deep neural networks (DNN) to adapt the data distribution in different domain and improved the previous works.
Wasserstein distance proposed by \cite{arjovsky2017wasserstein} can be used as a new distance measurement of domains to find better mapping.

\subsection{Network-based deep transfer learning}

\begin{figure}[h]
	\centering
	\setlength{\fboxsep}{0cm} 
	\includegraphics[page=3,trim=4.9cm 7.1cm 9cm 4.2cm,clip,width=0.7\textwidth]{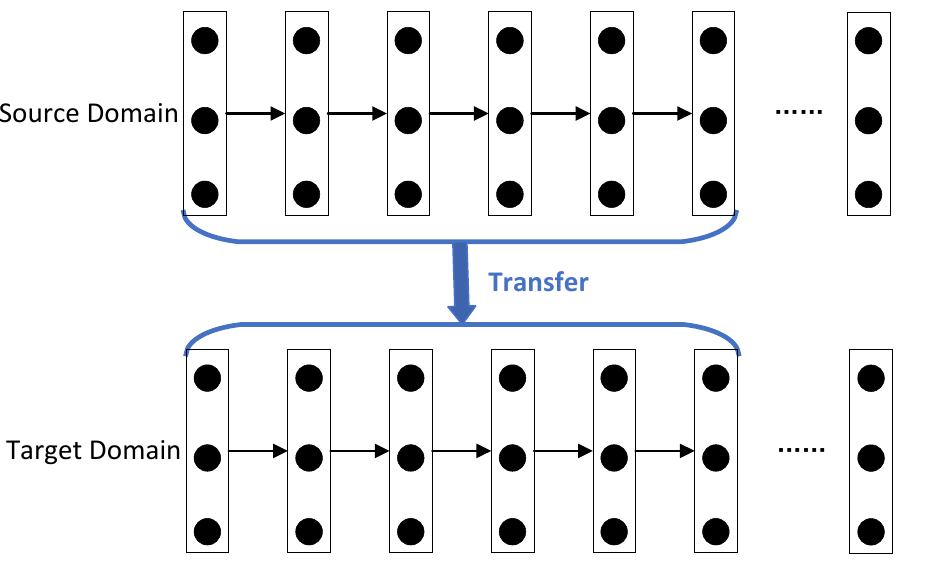}
	\caption{Sketch map of network-based deep transfer learning. 
        First, network was trained in source domain with large-scale training dataset. Second, partial of network pre-trained for source domain are transfer to be a part of new network designed for target domain. Finally, the transfered sub-network may be updated in fine-tune strategy.
        }
	\label{Processing of Network-based deep transfer learning}
\end{figure}

Network-based deep transfer learning refers to the reuse the partial network that pre-trained in the source domain, including its network structure and connection parameters, transfer it to be a part of deep neural network which used in target domain.
It is based on the assumption that "\textit{Neural network is similar to the processing mechanism of the human brain, and it is an iterative and continuous abstraction process. The front-layers of the network can be treated as a feature extractor, and the extracted features are versatile.}".
The sketch map of network-based deep transfer learning are shown in Fig. \ref{Processing of Network-based deep transfer learning}.

\cite{huang2013cross} divide the network into two parts, the former part is the language-independent feature transform and the last layer is the language-relative classifier. The language-independent feature transform can be transfer between multi languages.
\cite{oquab2014learning} reuse front-layers trained by CNN on the ImageNet dataset to compute intermediate image representation for images in other datasets, CNN are trained to learning image representations that can be efficiently transferred to other visual recognition tasks with limited amount of training data.
\cite{long2016unsupervised} proposed a approach to jointly learn adaptive classifiers and transferable features from labeled data in the source domain and unlabeled data in the target domain, which explicitly learn the residual function with reference to the target classifier by plugging several layers into deep network.
\cite{zhu2016deep} learning domain adaptation and deep hash features simultaneously in a DNN.
\cite{chang2017unsupervised} proposed a novel multi-scale convolutional sparse coding method. This method can automatically learns filter banks at different scales in a joint fashion with enforced scale-specificity of learned patterns, and provides an unsupervised solution for learning transferable base knowledge and fine-tuning it towards target tasks.
\cite{george2017deep} apply deep transfer learning to transfer knowledge from real-world object recognition tasks to glitch classifier for the detector of multiple gravitational wave signals. It demonstrate that DNN can be used as excellent feature extractors for unsupervised clustering methods to identify new classes based on their morphology, without any labeled examples.

Another very noteworthy result is that \cite{yosinski2014transferable} point out the relationship between network structure and transferability. It demonstrated that some modules may not influence in-domain accuracy but influence the transferability.
It point out what features are transferable in deep networks and which type of networks are more suitable for transfer. Given an conclusion that LeNet, AlexNet, VGG, Inception, ResNet are good chooses in network-based deep transfer learning.

\subsection{Adversarial-based deep transfer learning}

\begin{figure}[h]
	\centering
	\setlength{\fboxsep}{0cm} 
	\includegraphics[page=4,trim=4cm 10.2cm 9.2cm 3.4cm,clip,width=0.8\textwidth]{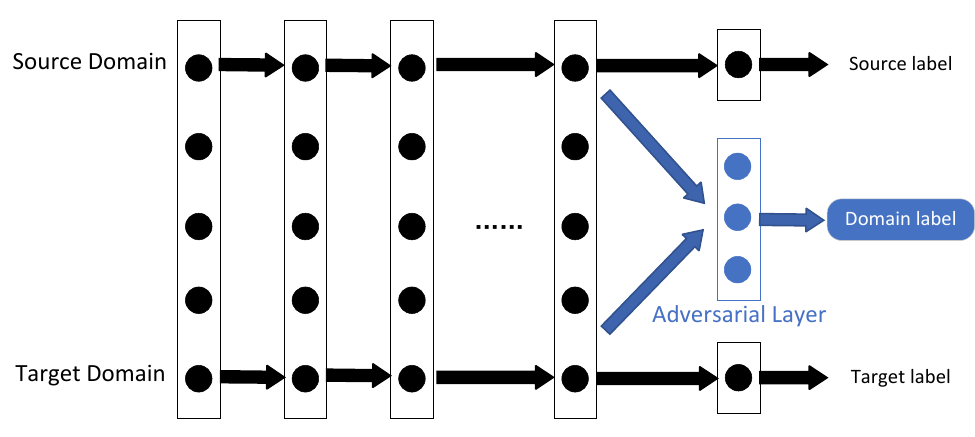}
	\caption{Sketch map of adversarial-based deep transfer learning. 
        In the training process on large-scale dataset in the source domain, the front-layers of network is regarded as a feature extractor. It extracting features from two domains and sent them to adversarial layer. The adversarial layer try to discriminates the origin of the features. If the adversarial network achieves worse performance, it means a small difference between the two types of feature and better transferability, and vice versa. 
        In the following training process, the performance of the adversarial layer will be considered to force the transfer network discover general features with more transferability.}
	\label{Processing of adversarial-based deep transfer learning}
\end{figure}

Adversarial-based deep transfer learning refers to introduce adversarial technology inspired by generative adversarial nets (GAN) \cite{goodfellow2014generative} to find transferable representations that is applicable to both the source domain and the target domain.
It is based on the assumption that "\textit{For effective transfer, good representation should be discriminative for the main learning task and indiscriminate between the source domain and target domain.}" 
The sketch map of adversarial-based deep transfer learning are shown in Fig. \ref{Processing of adversarial-based deep transfer learning}.

The adversarial-based deep transfer learning has obtained the flourishing development in recent years due to its good effect and strong practicality.
\cite{ajakan2014domain} introduce adversarial technology to transfer learning for domain adaption, by using a domain adaptation regularization term in the loss function.
\cite{ganin2014unsupervised} proposed an adversarial training method that suitable for most any feed-forward neural model by augmenting it with few standard layers and a simple new gradient reversal layer.
\cite{tzeng2015simultaneous} proposed a approach transfer knowledge cross-domain and cross-task simultaneity for sparsely labeled target domain data. A special joint loss function was used in this work to force CNN to optimize both the distance between domains which defined as $\mathcal{L}_D = \mathcal{L}_{c} + \lambda \mathcal{L}_{adver}$, 
where $\mathcal{L}_{c}$ is classification loss, $\mathcal{L}_{adver}$ is domain adversarial loss. Because the two losses stand in direct opposition to one another, an iterative optimize algorithm are introduced to update one loss when fixed another.
\cite{tzeng2017adversarial} proposed a new GAN loss and combine with discriminative modeling to a new domain adaptation method.
\cite{long2017domain} proposed a randomized multi-linear adversarial networks to exploit multiple feature layers and the classifier layer based on a randomized multi-linear adversary to enable both deep and discriminative adversarial adaptation.
\cite{luo2017label} utilize a domain adversarial loss, and generalizes the embedding to novel task using a metric learning-based approach to find more tractable features in deep transfer learning.

\section{Conclusion}

In this survey paper, we have review and category current researches of deep transfer learning. Deep transfer learning is classified into four categories for the first time: instances-based deep transfer learning, mapping-based deep transfer learning, network-based deep transfer learning, and adversarial-based deep transfer learning. In most practical applications, the above multiple technologies are often used in combination to achieve better results.
Most current researches focuses on supervised learning, how to transfer knowledge in unsupervised or semi-supervised learning by deep neural network may attract more and more attention in the future. 
Negative transfer and transferability measures are important issues in traditional transfer learning. The impact of these two issues in deep transfer learning also requires us to conduct further research.
In addition, a very attractive research area is to find a stronger physical support for transfer knowledge in deep neural network, which requires the cooperation of physicists, neuroscientists and computer scientists.
It can be predicted that deep transfer learning will be widely applied to solve many challenging problems with the development of deep neural network.

\bibliographystyle{splncs04}
\bibliography{icann2018_reference}

\end{document}